# Efficiency Enhancement of Genetic Algorithms via Building-Block-Wise Fitness Estimation

**Kumara Sastry**
**Martin Pelikan**
**David E. Goldberg**

# Efficiency Enhancement of Genetic Algorithms via Building-Block-Wise Fitness Estimation

Kumara Sastry
Illinois Genetic Algorithms Laboratory (IlliGAL), and
Department of Material Science & Engineering
University of Illinois at Urbana-Champaign
`ksastry@uiuc.edu`

Martin Pelikan
Department of Math & Computer Science
University of Missouri, St. Louis
`mpelikan@cs.umsl.edu`

David E. Goldberg
Illinois Genetic Algorithms Laboratory (IlliGAL), and
Department of General Engineering
University of Illinois at Urbana-Champaign
`deg@uiuc.edu`

**Abstract**

This paper studies fitness inheritance as an efficiency enhancement technique for a class of *competent* genetic algorithms called estimation distribution algorithms. Probabilistic models of important sub-solutions are developed to estimate the fitness of a proportion of individuals in the population, thereby avoiding computationally expensive function evaluations. The effect of fitness inheritance on the convergence time and population sizing are modeled and the speed-up obtained through inheritance is predicted. The results show that a fitness-inheritance mechanism which utilizes information on building-block fitnesses provides significant efficiency enhancement. For additively separable problems, fitness inheritance reduces the number of function evaluations to about half and yields a speed-up of about 1.75–2.25.

# 1 Introduction

Since the inception of genetic and evolutionary algorithms (GEAs), significant advances have been made in the theory, design and application to complex real-world problems. A decomposition methodology has been proposed for a successful design of GEAs (Goldberg, 1991; Goldberg, Deb, & Clark, 1992; Goldberg, 2002). Based on the design-decomposition theory several *competent* GEAs—genetic algorithms (GAs) that solve hard problems quickly, reliably, and accurately—have been proposed (Goldberg, 2002). Competent GEAs successfully solve boundedly difficult problems, oftentimes requiring polynomial—in terms of problem size—number of function evaluations. In essence, competent GEAs take problems that were intractable with first generation GAs, and render them *tractable*. One such class of competent GAs is the probabilistic model-building GAs (PMBGAs) (Pelikan, Lobo, & Goldberg, 2002; Larrañaga & Lozano, 2002; Pelikan, 2002), wherein a traditional crossover is replaced by a two step process: (1) Building of probabilistic models that identify important sub-solutions (or building blocks (BBs)). (2) Sampling the probabilistic models to efficiently mix the sub-solutions to create the offspring population. PMBGAs have solved problems of bounded difficulty (both regular as well as hierarchical) requiring only quadratic number of function evaluations (Pelikan, 2002; Pelikan, Goldberg, & Cantú-Paz, 2000b; Pelikan & Goldberg, 2001; Pelikan, Sastry, & Goldberg, 2003).

However, for large-scale problems, especially if the fitness function is a complex simulation, model, or computation, the task of computing even subquadratic number of evaluations can be daunting. This places a premium on a variety of *efficiency-enhancement techniques* (EETs). In essence, while competence leads us from intractability to tractability, efficiency enhancement takes us from tractability to *practicality*. Various EETs can be broadly classified into four broad classes (Goldberg, 2002): parallelization, hybridization, time-utilization and *evaluation relaxation*. In evaluation-relaxation schemes the computationally expensive, but accurate, fitness function is replaced by a cheap, but approximate, fitness function, there by speeding-up the GA process. One such evaluation-relaxation technique is *fitness inheritance* (Smith, Dike, & Stegmann, 1995; Sastry, Goldberg, & Pelikan, 2001). In fitness inheritance, a certain proportion of the offspring derive their fitness values from parental fitnesses rather than through function evaluations.

This paper analyzes fitness inheritance in PMBGAs, specifically extended compact genetic algorithm (eCGA) (Harik, 1999). In our study, the inherited fitness is derived and estimated from the probabilistic model constructed by eCGA. The purpose of this paper is to model such an inheritance mechanism and predict the scalability, speed-up and the optimal inheritance parameter that yields greatest speed-up. The approach used in this study is along the lines of those reported elsewhere for fitness inheritance in simple GAs (Sastry, Goldberg, & Pelikan, 2001). The analysis, in which the inherited fitness of an offspring was the average of its parental fitness values, predicted that using inherited fitness for 50% of the offspring population yielded a maximum speed-up of 30%. This was in contrast to a much more significant empirical speed-up observed by Smith, Dike, and Stegmann (Smith, Dike, & Stegmann, 1995). The purpose of this paper is also to investigate if an inheritance mechanism that incorporate knowledge of building block (or sub-solution) fitnesses would provide greater speed up than a simple inheritance procedure.

This paper is organized as follows. The next section reviews the literature on fitness inheritance in GAs, followed by a brief description of extended compact genetic algorithm in Section 3. We derive facetwise models for convergence time and population sizing required for the successful design of eCGA with fitness inheritance in Section 4. Subsequently, we use the convergence-time and population-sizing models to predict the scalability and speed-up obtained via fitness inheritance. Section 6 discusses possible directions of future research, followed by conclusions.

## 2 Literature Review

Smith, Dike and Stegmann (Smith, Dike, & Stegmann, 1995) proposed fitness inheritance in GAs. They proposed two ways of inheriting fitness, one by taking the average fitness and the other by taking a weighted average of the fitness of the two parents. Their results indicated that GAs with fitness inheritance outperformed those without inheritance in both the OneMax and an aircraft routing problem. However, they did not investigate the effect of fitness inheritance on convergence time, population sizing, and scalability of GAs. Though the original study showed very encouraging results, unfortunately there have been very few follow up studies on fitness inheritance. Zheng, Julstrom, and Cheng (Zheng, Julstrom, & Cheng, 1997) used fitness inheritance for the design of vector quantization codebooks.

Sastry, Goldberg, and Pelikan (Sastry, Goldberg, & Pelikan, 2001) provided theoretical analysis of fitness inheritance in simple GAs, in which facetwise models for convergence time and population sizing were derived and utilized for predicting the scalability and speed-up obtained by using fitness inheritance. The analysis predicted that an optimal speed-up of 30% can be achieved when 50% of the offspring population received inherited fitness. Chen, Goldberg, Ho, and Sastry (Chen, Goldberg, Ho, & Sastry, 2002) extended the analysis of fitness inheritance to multiobjective problems and predicted that a speed-up of 25% can be achieved when 50-55% of the offspring population received inherited fitness. Ducheyne, De Baets, and De Wulf (Ducheyne, De Baets, & De Wulf, 2003) studied the utility of fitness inheritance in convex and non-convex multiobjective problems. Pelikan and Sastry (Pelikan & Sastry, 2004) have proposed fitness inheritance for the efficiency-enhancement of Bayesian optimization algorithm.

## 3 Extended Compact Genetic Algorithm (eCGA)

The extended compact GA proposed by Harik (Harik, 1999) is based on a key idea that the choice of a good probability distribution is equivalent to linkage learning. The measure of a good distribution is quantified based on minimum description length (MDL) models. The key concept behind MDL models is that all things being equal, simpler distributions are better than more complex ones. The MDL restriction penalizes both inaccurate and complex models, thereby leading to an optimal probability distribution. Thus, MDL restriction reformulates the problem of finding a good distribution as an optimization problem that minimizes both the probability model as well as population representation. The probability distribution used in eCGA is a class of probability models known as marginal product models (MPMs). MPMs are formed as a product of marginal distributions on a partition of the genes and are similar to those of the compact GA (CGA) (Harik, Lobo, & Goldberg, 1998) and PBIL (Baluja, 1994). Unlike the models used in CGA and PBIL, MPMs can represent probability distributions for more than one gene at a time. MPMs also facilitate a direct linkage map with each partition separating tightly linked genes. For example, the following MPM, `[1,3][2][4]`, for a four-bit problem represents that the $1^{st}$ and $3^{rd}$ genes are linked and $2^{nd}$ and $4^{th}$ genes are independent. Additionally, the MPM consists of the marginal probabilities: $\{p(x_1 = 0, x_3 = 0), p(x_1 = 0, x_3 = 1), p(x_1 = 1, x_3 = 0), p(x_1 = 1, x_3 = 1), p(x_2 = 0), p(x_2 = 1), p(x_4 = 0), p(x_4 = 1)\}$, where $x_i$ is the value of the $i^{th}$ gene.

The eCGA can be algorithmically outlined as follows:

1. Initialization: The population is usually initialized with random individuals. However, other initialization procedures can also be used.

2. Evaluate the fitness value of the individuals

3. Selection: The eCGA uses s-wise tournament selection (Goldberg, Korb, & Deb, 1989). However, other selection procedures can be used instead of tournament selection.

4. Build the probabilistic model: In eCGA, both the structure and the parameters of the model

are searched. A greedy search heuristic is used to find an optimal model of the selected individuals in the population.

5. Create new individuals: In eCGA, new individuals are created by sampling the probabilistic model.

6. Replace the parents with the offspring.

7. Repeat steps 2–6 until some termination criteria are met.

Two things need further explanation, one is the identification of MPM using MDL and the other is the creation of a new population based on MPM. The identification of MPM is formulated as a constrained optimization problem,

$$\text{Minimize} \quad C_m + C_p \tag{1}$$
$$\text{Subject to}$$
$$2^{k_i} \leq n \quad \forall i \in [1, m]. \tag{2}$$

where $C_m$ is the model complexity which represents the cost of a complex model. In essence, the model complexity, $C_m$, quantifies the model representation size in terms of number of bits required to store all the marginal probabilities. Let, a given problem of size $\ell$ with binary alphabets, have $m$ partitions with $k_i$ genes in the $i^{\text{th}}$ partition, such that $\sum_{i=1}^{m} k_i = \ell$. Then each partition $i$ requires $2^k - 1$ independent frequencies to completely define its marginal distribution. Furthermore, each frequency is of size $\log_2(n)$, where $n$ is the population size. Therefore, the model complexity $C_m$, is given by

$$C_m = \log_2(n) \sum_{i=1}^{m} \left(2^{k_i} - 1\right). \tag{3}$$

The compressed population complexity, $C_p$, represents the cost of using a simple model as against a complex one. In essence, the compressed population complexity, $C_p$, quantifies the data compression in terms of the entropy of the marginal distribution over all partitions. Therefore, $C_p$ is evaluated as

$$C_p = n \sum_{i=1}^{m} \sum_{j=1}^{2^{k_i}} -p_{ij} \log_2(p_{ij}) \tag{4}$$

where $p_{ij}$ is the frequency of the $j^{\text{th}}$ gene sequence of the genes belonging to the $i^{\text{th}}$ partition. In other words, $p_{ij} = N_{ij}/n$, where $N_{ij}$ is the number of chromosomes in the population (after selection) possessing bit-sequence $j \in [1, 2^{k_i}]$ [1] for $i^{\text{th}}$ partition. The constraint (Equation 2) arises due to finite population size.

The following greedy search heuristic is used to find an optimal or near-optimal probabilistic model:

1. Assume each variable is independent of each other. The model is a vector of probabilities.

2. Compute the model complexity and population complexity values of the current model.

---

[1] Note that a BB of length $k$ has $2^k$ possible sequences where the first sequence denotes be $00\cdots 0$ and the last sequence $11\cdots 1$

3. Consider all possible $\frac{1}{2}\ell(\ell-1)$ merges of two variables.

4. Evaluate the model and compressed population complexity values for each model structure.

5. Select the merged model with lowest combined complexity.

6. If the combined complexity of the best merged model is better than the combined complexity of the model evaluated in step 2., replace it with the best merged model and go to step 2.

7. If the combined complexity of the best merged model is less than or equal to the combined complexity, the model cannot be improved and the model of step 2. is the probabilistic model of the current generation.

The offspring population are generated by randomly choosing subsets from the current individuals according to the probabilities of the subsets as calculated in the probabilistic model.

## 4 Fitness Inheritance in eCGA

The previous section outlined the key steps and mechanisms of eCGA. In this section we discuss the enhancements and modifications made on eCGA to enable the incorporation of fitness inheritance.

Similar to earlier studies on fitness inheritance, all the individuals in the initial population are evaluated and subsequently a portion of the offspring population receives inherited fitness and the others receive actual fitness evaluation. That is, an offspring receives inherited fitness with a probability $p_i$, or an evaluated fitness with a probability $1 - p_i$. However, unlike previous works, which used either average or weighted average of parental fitnesses as the inherited fitness, here we employ the probabilistic model built by eCGA and estimates of linkage-group fitnesses in determining the inherited fitness. Specifically, individuals from parental population that received evaluated fitnesses (that is, individuals whose fitnesses were not inherited) are used to determine the fitnesses of schemata that are defined by the probabilistic model. The schema fitness from different partitions are then used to estimate the fitness of an offspring. The fitness-inheritance procedure is detailed in the following paragraphs.

After the probabilistic model is built and the linkage map is obtained (step 4 of the eCGA algorithm outlined in the previous section), we estimate the fitness of schemata using only those individuals whose fitnesses were not inherited. In all, we estimate the fitness of a total of $\sum_{i=1}^{m} 2^{k_i}$ schemas. Considering our previous example (Section 3) of a four-bit problem, whose model is [1,3][2][4], the schemata whose fitnesses are estimated are: {0*0*, 0*1*, 1*0*, 1*1*, *0**, *1**, ***0, ***1}.

The fitness of a schema, $h$, is defined as the difference between the average fitness of individuals that contain the schema and the average fitness of all the individuals. That is,

$$\hat{f}_s(h) = \frac{1}{n_h} \sum_{\{i|x_i \supset h\}} f(x_i) - \frac{1}{n'} \sum_{i=1}^{n'} f(x_i) \qquad (5)$$

where $n_h$ is the total number of individuals that contain the schema $h$, $x_i$ is the $i^{\text{th}}$ individual and $f(x_i)$ is its fitness, $n'$ is the total number of individuals that were evaluated. If a particular schema is not present in the population, its fitness is arbitrarily set to zero. Furthermore, it should be noted that the above definition of schema fitness is not unique and other estimates can be used. The key point however is the use of the probabilistic model in determining the schema fitnesses.

Once the schema fitnesses across partitions are estimated, the offspring population is created as outlined in Section 3. An offspring receives inherited fitness with a probability $p_i$, referred to as the inheritance probability. The inherited fitness is computed as follows:

$$f_{\text{inh}}(y) = \frac{1}{n'} \sum_{i=1}^{n'} f(x_i) + \sum_{i=1}^{m} \hat{f}_s(h_i \in y) \tag{6}$$

where $y$ is the offspring individual. It should be noted that the eCGA model yields non-overlapping linkage groups and might not be appropriate for problems with overlapping BBs. However, similar concepts can be incorporated in other PMBGAs such as the Bayesian optimization algorithm (BOA) (Pelikan, Goldberg, & Cantú-Paz, 2000b) which can handle overlapping BBs. Moreover, the inherited fitness can be computed by other methods, some of which are outlined in (Pelikan & Sastry, 2004), but the key is to use the estimates of substructure fitnesses in the computation.

With this understanding of the inheritance mechanism, we will now model the effects of fitness inheritance on population sizing and convergence time in the next sections. These models are then used to predict the speed-up (or efficiency enhancement) obtained through fitness inheritance.

## 5 Modeling Fitness Inheritance

To ease the analytical burden, we assume a non-overlapping population of fixed size and a generationwise eCGA. We consider binary strings of fixed length as the chromosomes. Furthermore, the models assume additively separable (nearly separable) uniformly-scaled problems. That is, we assume that the BBs are non-overlapping and the contribution of each BB to fitness is equal. Many of the above assumptions are made for simplifying the models and can be relaxed in a straightforward manner. We also assume that the probabilistic model accurately map to the building blocks of the problem. However, the effect of a wrong model is not significant as we will see later in the verification of the models.

Assuming that the actual fitness distribution, F, is Gaussian with mean $\mu_f$ and variance $\sigma_{f,t}^2$.

$$F = \mathcal{N}\left(\mu_{f,t}, \sigma_{f,t}^2\right),$$

and that the distribution of fitness with inheritance, F' is Gaussian with mean $\mu_{f',t}$ and variance $\sigma_{f',t}^2$.

$$F' = \mathcal{N}\left(\mu_{f',t}, \sigma_{f',t}^2\right).$$

The above assumptions are reasonable since crossover has a normalizing effect.

From (Sastry, Goldberg, & Pelikan, 2001) we know that fitness inheritance can be modeled as an addition of external noise to the actual fitness.

$$F' = F + \mathcal{N}\left(0, p_i \sigma_{i,t}^2\right) \tag{7}$$

where $p_i$ is the probability of and individual receiving inherited fitness, and $\sigma_{i,t}^2$ is variance of error between the inherited and actual fitnesses. Since the inherited fitness, $f_{\text{inh}}$, is equal to the sum of estimated building-block fitnesses, for additively separable uniformly scaled problems $\sigma_{i,t}^2$ is given by

$$\begin{aligned}
\sigma_{i,t}^2 &= \sum_{j=1}^{m} \sigma^2\left(\hat{f}_{BB_j} - f_{BB_j}\right), \\
&= m\sigma_{BB,\text{inh}}^2,
\end{aligned} \quad (8)$$

where $\sigma_{BB,\text{inh}}^2$ is the variance of the estimated fitness of a building block. Since the noise in the estimate of a BB comes from the other $m-1$ partitions (Goldberg, Deb, & Clark, 1992),

$$\sigma_{BB,\text{inh}}^2 \approx [(m-1)/m]\sigma_{BB}^2 \approx \sigma_{BB}^2.$$

where $\sigma_{BB}^2$ is the actual BB variance. This approximation however is not valid for very high inheritance-probability values as the BB fitness is estimated from very few individuals which increase the error in the estimate significantly. Empirically, we observed that $\sigma_{BB,\text{inh}}^2$ becomes significantly higher than the above approximation when $p_i \geq 0.85$.

Therefore, the noise variance due to fitness inheritance is given by

$$\sigma_{i,t}^2 = m\sigma_{BB}^2 = \sigma_{f,t}^2. \quad (9)$$

From the above equation and Equations 5 and 7, we can write the fitness distribution with inheritance as

$$F' = \mathcal{N}\left(\mu_{f,t}, (1+p_i)\,\sigma_{f',t}^2\right). \quad (10)$$

We now proceed to develop population-sizing and convergence-time models.

## 5.1 Population Sizing

Population sizing is one of the important factors that determine GA success. Therefore it is essential to appropriately size the population to incorporate the effects of fitness inheritance. Goldberg, Deb, and Clark (Goldberg, Deb, & Clark, 1992) proposed population-sizing models for correctly deciding between competing BBs. They incorporated noise arising from other partitions into their model. However, they assumed that if wrong BBs were chosen in the first generation, the GAs would be unable to recover from the error. Harik, Cantú-Paz, Goldberg, and Miller (Harik, Cantú-Paz, Goldberg, & Miller, 1999) refined the above model by incorporating cumulative effects of decision making over time rather than in first generation only. Harik et al. (Harik, Cantú-Paz, Goldberg, & Miller, 1997) modeled the decision making between the best and second best BBs in a partition as a gambler's ruin problem. This model is based on the assumption that the selection process used is tournament selection without replacement. Miller (Miller, 1997) extended the gambler's ruin model to incorporate external noise. Pelikan, Goldberg, and Cantú-Paz (Pelikan, Goldberg, & Cantú-Paz, 2000a) and Pelikan, Sastry, and Goldberg (Pelikan, Sastry, & Goldberg, 2003) developed population-sizing models for PMBGAs, specifically for BOA. Sastry and Goldberg (Sastry & Goldberg, 2000) empirically demonstrated that the population sizing of eCGA is similar to that of BOA.

The population-sizing model which incorporates the effect of model-building and its accuracy on the population sizing of the GA, is given by

$$n = -c_n \log(\alpha) 2^k \left(\sigma_f^2 + \sigma_N^2\right),$$

where $n$ is the population size, $c_n$ is a problem-dependent constant, $k$ is the BB length, $\alpha$ is the probability of failure, and $\sigma_f^2$ is the fitness variance, and $\sigma_N^2$ is the external noise variance. From Equation 10, we can see that $\sigma_N^2 = p_i \sigma_f^2$. Therefore, the population-size equation can be written as

$$n = -c_n \log(\alpha) 2^k \sigma_f^2 (1 + p_i). \tag{11}$$

The effect of fitness inheritance on population sizing can be predicted by dividing the above equation by the population size required when no inheritance is used. That is, by the population-sizing ratio, $n/n(p_i = 0)$:

$$n_r = \frac{n}{n(p_i = 0)} = (1 + p_i). \tag{12}$$

## 5.2 Time to Convergence

Understanding convergence time is one of the key factors in the design of GAs and in predicting their scalability (Goldberg, 2002). Mühlenbein and Schlierkamp-Voosen (Mühlenbein & Schlierkamp-Voosen, 1993) derived the convergence-time model for the breeder GA using the notion of *selection intensity* (Bulmer, 1985) from population genetics. Thierens and Goldberg (Thierens & Goldberg, 1994) derived convergence-time models for different selections schemes including binary tournament selection. Bäck (Bäck, 1994) derived estimates of selection intensity for $s$-wise tournament and $(\lambda, \mu)$ selection. Miller and Goldberg (Miller & Goldberg, 1995) developed convergence-time models for $s$-wise tournament selection and incorporated the effects of external noise. Bäck (Bäck, 1995) developed convergence-time models for $(\lambda, \mu)$ selection.

The assumptions used in developing the convergence-time model of Miller and Goldberg (Miller & Goldberg, 199 are applicable to PMBGAs and the model is therefore valid for BOA and eCGA as well (Pelikan, Goldberg, & Cantú An approximate form of the convergence-time model for noisy function evaluations can be written as (Goldberg, 2002):

$$t_c = c_t \sqrt{m \cdot k} \sqrt{1 + \frac{\sigma_N^2}{\sigma_f^2}}. \tag{13}$$

where $c_t$ is a problem dependent constant. A detailed derivation of the above equation and other approximations are given elsewhere (Sastry & Goldberg, 2002; Sastry, 2001).

Again the effect of fitness inheritance on the convergence time can be predicted by dividing the above equation by the convergence time when no inheritance is used. Recognizing that $\sigma_N^2 = p_i \sigma_f^2$, the convergence-time ratio, $t_c/t_c(p_i = 0)$ is given by

$$t_{c,r} = \frac{t_c}{t_c(p_i = 0)} = \sqrt{1 + p_i}. \tag{14}$$

## 5.3 Speed-Up

We now use the convergence-time and population-sizing models in the previous sections to predict the number of function evaluations required for eCGA success. The total number of function evaluations can be written as

$$n_{fe} = n + n(t_c - 1)(1 - p_i). \tag{15}$$

Recall that all the individuals in the initial population are evaluated and there after on an average $n(1 - p_i)$ individuals are evaluated.

To isolate the effect of fitness inheritance on the scalability of eCGA, we consider the ratio of total number of function evaluations required with fitness inheritance and that required without fitness inheritance. That is, we consider the function-evaluation ratio,

$$\begin{aligned} n_{fe,r} &= \frac{n_{fe}}{n_{fe}(p_i = 0)} = \frac{n + n\,(t_c - 1)\,(1 - p_i)}{n(p_i = 0)t_c(p_i = 0)}, \\ &= n_r\left[t_{c,r}\,(1 - p_i) + \frac{p_i}{t_c(p_i = 0)}\right], \quad (16) \\ n_{fe,r} &\approx n_r \cdot t_{c,r}\,(1 - p_i). \quad (17) \end{aligned}$$

Substituting Equations 12 and 14, in the above equation and simplifying, we get

$$n_{fe,r} \approx (1 + p_i)^{1.5}\,(1 - p_i) \quad (18)$$

The speed-up that can be obtained through fitness inheritance is given by the inverse of the function-evaluation ratio:

$$\eta_{\text{inh}} = \frac{1}{(1 + p_i)^{1.5}\,(1 - p_i)}. \quad (19)$$

Equation 18 indicates that the function-evaluation ratio increases (or the speed-up reduces) at low $p_i$ values, reaches a maximum at $p_i = 0.2$. When $p_i = 0.2$ the number of function evaluations required is 5% more than that required without inheritance. In other words, the speed-up at $p_i = 0.2$ is 0.95. For inheritance probabilities above 0.2 the function-evaluation ratio decreases (speed-up increases) with the inheritance probability. Equation 19 predicts that the speed-up is maximum when $p_i = 1.0$, however, it should be noted that the models derived are not entirely valid for higher $p_i$ values ($p_i \geq 0.95$). Empirically we observed that even when $p_i = 0.95$, we obtained a speed-up of 1.8–2.0. However, when $p_i = 1.0$, the number of function evaluations required were four times than that required without inheritance (speed-up of 0.25). The optimal inheritance probability therefore should be around the value where the model prediction deviates.

The models developed in these sections are empirically verified with the help of two test functions in the next section.

## 5.4 Model Verification

Before discussing the empirical verification of the models developed in the previous sections, we briefly describe the two test functions used for the model verification. Our approach in verifying the models and observing if fitness inheritance yields speed-up is to consider bounding *adversarial problems* that exploit one or more dimensions of problem difficulty (Goldberg, 2002). Particularly, we are interested in problems where building-block identification is critical for the GA success. Additionally, the problem solver (eCGA) should not have any knowledge of the BB structure of the problem, but should be known to researchers for verification.

The two test functions with the above properties and used in this study are:

1. **OneMax problem**, which is a GA-easy problem and in which each variable is independent of the others. While the optimization of the OneMax problem is easy, the probabilistic models built by eCGA (or PMBGAs) for OneMax, however, are known to be only partially correct and include spurious linkages (Sastry & Goldberg, 2000; Pelikan, Goldberg, & Sastry, 2001). Therefore, the inheritance results on the OneMax problem will indicate if the effect of using

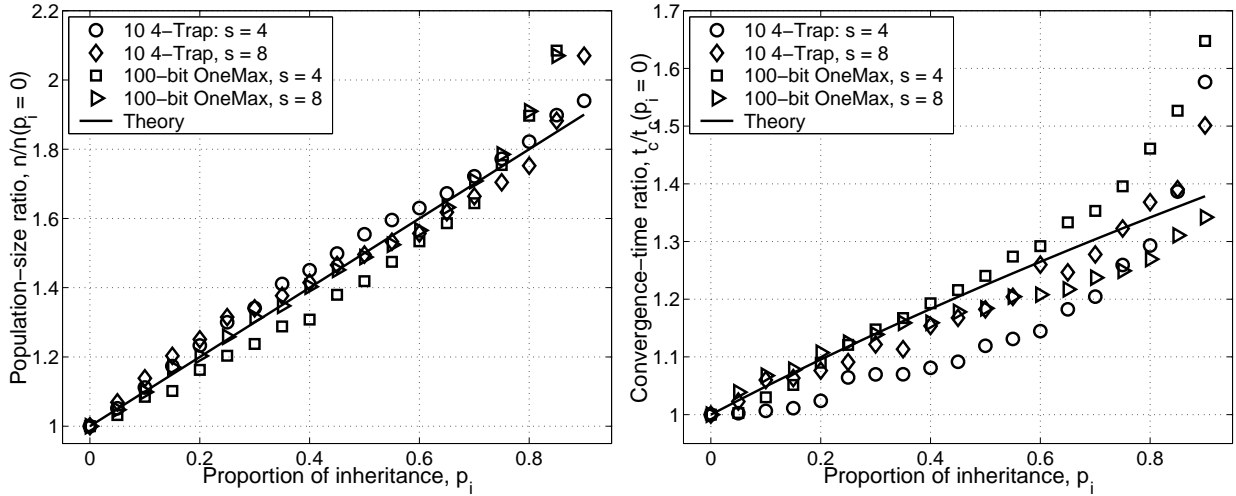

(a) Verification of population-size-ratio model   (b) Verification of convergence-time-ratio model

Figure 1: Verification of the population-size-ratio model (Equation 11) and convergence-time-ratio model (Equation 14) for various inheritance proportions with empirical results for 100-bit OneMax and 10 4-Trap problems. The population size is determined by a bisection method such that the failure probability averaged over 30–100 independent runs is $1/m$ (that is, $\alpha = 1/m$). The convergence-time is determined by the number of generations required to achieve convergence on $m - 1$ out of $m$ BBs correctly. The results are averaged over 30 independent bisection runs.

  partially correct linkage mapping on the inherited fitness is significant. We use a 100-bit OneMax problem for verifying the convergence-time and population-sizing models.

2. **m k-Deceptive *trap* problem**, which consists of additively separable *deceptive* functions (Goldberg, 1987; Deb & Goldberg, 1993; Deb & Goldberg, 1994). Deceptive functions are designed to thwart the very mechanism of selectorecombinative search by punishing any localized hillclimbing and requiring mixing of whole building blocks at or above the order of deception. Using such *adversarially* designed functions is a stiff test—in some sense the stiffest test—of algorithm performance. The idea is that if an algorithm can beat an adversarially designed test function, it can solve other problems that are equally hard or easier than the adversary.

We use a tournament selection with tournament sizes of 4 and 8 in obtaining the empirical results. An eCGA run is terminated when all the individuals in the population converge to the same fitness value. The average number of BBs correctly converged are computed over 30–100 independent runs. The minimum population size required such that $m - 1$ BBs converge to the correct value is determined by a bisection method (Sastry, 2001). The results of population-size and convergence-time ratio is averaged over 30 such bisection runs, while the results for the function-evaluation ratio is averaged over 900–3000 independent runs.

The population-size-ratio model (Equation 12) is verified with empirical results for OneMax and m k-Trap in Figure 1(a). The standard deviation for the empirical runs are very small ($\sigma \in$

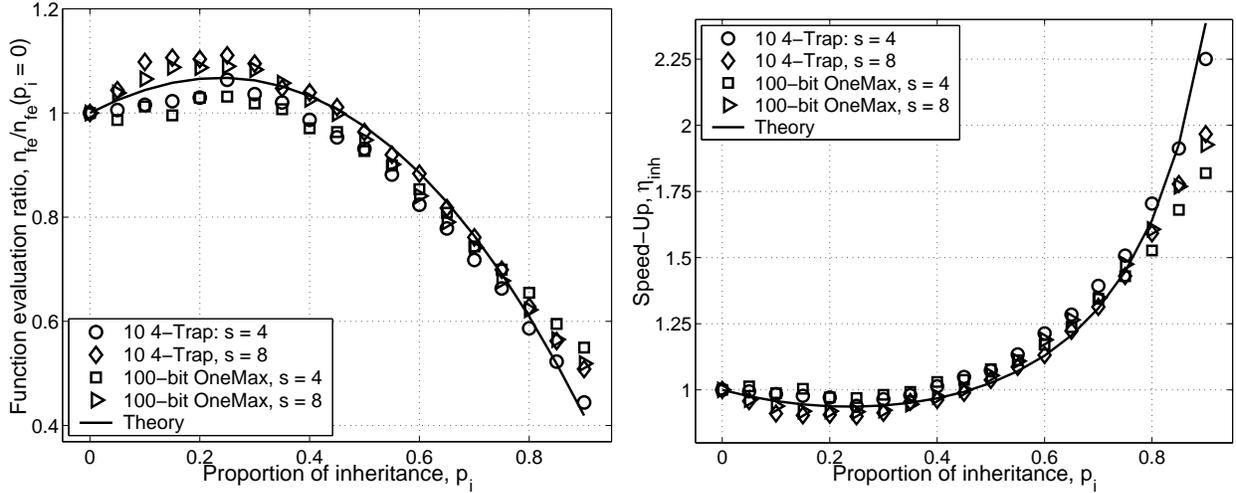

(a) Verification of function-evaluation-ratio model

(b) Verification of speed-up model

Figure 2: Verification of the function-evaluation-ratio model (Equation 18) and the speed-up model (Equation 19) with empirical results on 100-bit OneMax and 10 4-Trap problems. The total number of function evaluations is determined such that the failure probability of an eCGA run is at most $1/m$. The results are averaged over 900–3000 independent runs.

$[4 \times 10^{-4}, 1.8 \times^{-2}])$, and therefore the error bars are not shown in Figure 1(a). As shown in the figure, the empirical results agrees with the model. The population size required to ensure that, on an average, eCGA fails to converge on at most one out of $m$ BBs, increases linearly with the inheritance probability, $p_i$. The population sizes required at very high inheritance-probability values, $p_i \geq 0.85$ deviates from the predicted values. This is because the noise introduced due to inheritance increases significantly at higher $p_i$ values because of limited number of individuals with evaluated fitness that take part in the estimate of schemata fitnesses.

The verification of the convergence-time-ratio model (Equation 14 with empirical results for OneMax and m k-Trap are shown in Figure 1(b). The standard deviation for the empirical runs are very small ($\sigma \in [2 \times 10^{-4}, 2.7 \times 10^{-2}]$), and therefore the error bars are not shown. As shown in the figure, the agreement between the empirical results and the model are slightly poor when compared to that for population-size ratio. This is because of the approximations used in deriving the convergence time model and more accurate, but complex, models exist that improve the predictions (Sastry, 2001). However, as we will see later, this disagreement between the model and experiments does not significantly affect the prediction of speed-up, which is the key objective. The empirical convergence-time ratio deviates from the predicted value at slightly lower inheritance probabilities, $p_i \geq 0.75$ than the population-size ratio. This is to be expected as the population sizing is largely dictated by the fitness and noise variances in the initial few generations, while the convergence time is dictated by the fitness and noise variances over the GA run. Therefore, the effect of high $p_i$ values, or less number of evaluated individuals, is cumulative over time and leads to deviation from theory at lower $p_i$ values than the population size.

The predicted values of function-evaluation-ratio (Equation 18) and the speed-up (Equation 19) are verified with empirical results for OneMax and m k-Trap in Figures 2(a) and 2(b), respectively. The standard deviation for the empirical runs are very small ($\sigma \in [7 \times 10^{-5}, 7 \times 10^{-3}]$), and therefore are not shown. As shown in Figures 2(a) and 2(b), the empirical results agree with the analytical models. Furthermore, the agreement for the OneMax problem with the models is good even though the building-block identification for the OneMax problem is only partially correct. The results show that the required number of function evaluations is almost of halved with the use of fitness inheritance thereby leading to a speed-up of 1.75–2.25. This is a significant improvement over a speed-up of 1.3 observed for simple GAs with a simple inheritance mechanism (Sastry, Goldberg, & Pelikan, 2001). Furthermore, fitness inheritance yields speed-up even when the inheritance probability is very high (as high as 0.95), which agrees with the empirical observation of Smith, Dike, and Stegmann (Smith, Dike, & Stegmann, 1995).

Overall, the results suggest that significant efficiency enhancement can be achieved through an inheritance mechanism that incorporates knowledge of important sub-solutions of a problem and their partial fitnesses.

## 6  Future Work

In this paper, we have introduced and analyzed a fitness inheritance mechanism that incorporates the knowledge of building-block structure and fitness. The BB structure and fitnesses are identified and estimated with the help of extended compact genetic algorithm, a competent GA. The results show that significant speed-up can be obtained through fitness inheritance and warrants further research and enhancements in many avenues some of which are listed in the following:

- **Problems with overlapping building blocks:** While this paper considered problems with non-overlapping building blocks, many problems have different building blocks that share common components. While considering problems with overlapping building blocks, eCGA might not be the appropriate search method, however the fitness inheritance mechanism should still be valid which can be used in other more advanced PMBGAs such as the Bayesian optimization algorithm (Pelikan & Sastry, 2004).

- **Non-Uniformly scaled problems:** In this paper, all the building blocks of a problem had equal salience, which might not be the case in real-world problems. The non-uniform scaling induces sequentiality in the identification and convergence of the building blocks. The effect of non-uniform building-block salience on the speed-up and optimal inheritance proportion should be investigated.

- **Hierarchical problems:** One of the important class of nearly decomposable problems is hierarchical problems (Pelikan & Goldberg, 2001), in which the building-block interactions are present at more than a single level. Such problems can be successfully solved in polynomial time by the hierarchical Bayesian optimization algorithm . The fitness inheritance mechanism used in this study could be enhanced and incorporated into hBOA and the efficiency enhancement provided by inheritance can be investigated.

- **Additional dimensions of problem difficulty:** In this paper we considered one of the dimensions of GA problem difficulty, deception. However, there are other dimensions of problem difficulty (Goldberg, 2002) such as epistasis and external noise. This factors should be included in isolation or in conjunction with other factors of problem difficulty in determining a complete picture of efficiency enhancement provided by fitness inheritance.

- **Real-World problems:** One of the key objectives of analyzing and developing fitness-inheritance mechanism is to aid the principled design of such a mechanism in competent genetic algorithms for successfully solving complex real-world problems in *practical* time.

- **Interactive and Human-Based evolutionary algorithms:** In interactive evolutionary algorithms the fitness of a solution is given by a human being rather than by a computation (Takagi, 2001). In human-based GAs, both fitness evaluations and genetic operations such as selection, crossover, and mutation are performed by users (Kosorukoff & Goldberg, 2002). To avoid overwhelming the users, it is often infeasible to use large populations even if the search problem might warrant one. Fitness inheritance can be used in such cases to alleviate this restriction. While only a small portion of the population gets the fitness from the users, the rest can receive inherited fitness, thereby allowing larger population sizes but still maintaining the same number of function evaluations from the users.

# 7 Conclusions

In this paper, we introduced a fitness inheritance mechanism that estimates the fitness by schema fitness. The sub-solutions (or schemata) are automatically and adaptively identified by a probabilistic model building genetic algorithm called extended compact genetic algorithm. We have also developed a theoretical basis for fitness inheritance and have derived models for convergence time and population sizing. The convergence-time and population-sizing models were in turn used to predict the effect of fitness inheritance on the scalability of eCGA and also to predict the speed-up obtained via fitness inheritance. We observed that the fitness inheritance mechanism that incorporates information of building-block structure and fitness provides a significant speed-up. For additively separable problems, the results show that using evaluations for only about 5-15% of the population, fitness inheritance can yield a speed-up of around 1.75–2.25.

## acknowledgments

We gratefully acknowledge Robert E. Smith for his helpful suggestions. We also thank Martin Butz, Ying-ping Chen, Xavier Llora, Kei Ohnishi, and Tian-Li Yu for many useful discussions.

This work was sponsored by the Air Force Office of Scientific Research, Air Force Materiel Command, USAF, under grant F49620-00-0163 and F49620-03-1-0129, the National Science Foundation under ITR grant DMR-99-76550 (at Materials Computation Center), and ITR grant DMR-0121695 (at CPSD), and the Dept. of Energy under grant DEFG02-91ER45439 (at Fredrick Seitz MRL). The U.S. Government is authorized to reproduce and distribute reprints for government purposes notwithstanding any copyright notation thereon.

The views and conclusions contained herein are those of the authors and should not be interpreted as necessarily representing the official policies or endorsements, either expressed or implied, of the Air Force Office of Scientific Research, the National Science Foundation, or the U.S. Government.

# References

Bäck, T. (1994). Selective pressure in evolutionary algorithms: A characterization of selection mechanisms. *Proceedings of the First IEEE Conference on Evolutionary Computation*, 57–62.

Bäck, T. (1995). Generalized convergence models for tournament—and $(\mu, \lambda)$—selection. *Proceedings of the Sixth International Conference on Genetic Algorithms*, 2–8.

Baluja, S. (1994). *Population-based incremental learning: A method of integrating genetic search based function optimization and competitive learning* (Technical Report CMU-CS-94-163). Carnegie Mellon University.

Bulmer, M. G. (1985). *The mathematical theory of quantitative genetics*. Oxford: Oxford University Press.

Chen, J.-H., Goldberg, D. E., Ho, S.-Y., & Sastry, K. (2002). Fitness inheritance in multi-objective optimization. *Proceedings of the Genetic and Evolutionary Computation Conference*, 319–326. (Also IlliGAL Report No. 2002017).

Deb, K., & Goldberg, D. E. (1993). Analyzing deception in trap functions. *Foundations of Genetic Algorithms*, *2*, 93–108. (Also IlliGAL Report No. 91009).

Deb, K., & Goldberg, D. E. (1994). Sufficient conditions for deceptive and easy binary functions. *Annals of Mathematics and Artificial Intelligence*, *10*, 385–408. (Also IlliGAL Report No. 92001).

Ducheyne, E., De Baets, B., & De Wulf, R. (2003). Is fitness inheritance useful for real-world applications? *Proceedings of the Evolutionary Multi-Objective Conference*, 31–42.

Goldberg, D. E. (1987). Simple genetic algorithms and the minimal, deceptive problem. In Davis, L. (Ed.), *Genetic algorithms and simulated annealing* (Chapter 6, pp. 74–88). Los Altos, CA: Morgan Kaufmann.

Goldberg, D. E. (1991). Theory tutorial. (Tutorial presented with G. Liepens at the 1991 International Conference on Genetic Algorithms, La Jolla, CA).

Goldberg, D. E. (2002). *Design of innovation: Lessons from and for competent genetic algorithms*. Boston, MA: Kluwer Acadamic Publishers.

Goldberg, D. E., Deb, K., & Clark, J. H. (1992). Genetic algorithms, noise, and the sizing of populations. *Complex Systems*, *6*, 333–362. (Also IlliGAL Report No. 91010).

Goldberg, D. E., Korb, B., & Deb, K. (1989). Messy genetic algorithms: Motivation, analysis, and first results. *Complex Systems*, *3*(5), 493–530. (Also IlliGAL Report No. 89003).

Harik, G. (1999, January). *Linkage learning via probabilistic modeling in the ECGA* (IlliGAL Report No. 99010). Urbana, IL: University of Illinois at Urbana-Champaign.

Harik, G., Cantú-Paz, E., Goldberg, D. E., & Miller, B. L. (1997). The gambler's ruin problem, genetic algorithms, and the sizing of populations. *Proceedings of the IEEE International Conference on Evolutionary Computation*, 7–12. (Also IlliGAL Report No. 96004).

Harik, G., Cantú-Paz, E., Goldberg, D. E., & Miller, B. L. (1999). The gambler's ruin problem, genetic algorithms, and the sizing of populations. *Evolutionary Computation*, *7*(3), 231–253. (Also IlliGAL Report No. 96004).

Harik, G., Lobo, F., & Goldberg, D. E. (1998). The compact genetic algorithm. *Proceedings of the IEEE International Conference on Evolutionary Computation*, 523–528. (Also IlliGAL Report No. 97006).

Kosorukoff, A., & Goldberg, D. (2002). Evolutionary computation as a form of organization. *Proceedings of the Genetic and Evolutionary Computation Conference*, 965–972. (Also IlliGAL Report No. 2001004).

Larrañaga, P., & Lozano, J. A. (Eds.) (2002). *Estimation of distribution algorithms*. Boston, MA: Kluwer Academic Publishers.

Miller, B. L. (1997, May). *Noise, sampling, and efficient genetic algorithms*. Doctoral dissertation, University of Illinois at Urbana-Champaign, General Engineering Department, Urbana, IL. (Also IlliGAL Report No. 97001).

Miller, B. L., & Goldberg, D. E. (1995). Genetic algorithms, tournament selection, and the effects of noise. *Complex Systems*, *9*(3), 193–212. (Also IlliGAL Report No. 95006).

Mühlenbein, H., & Schlierkamp-Voosen, D. (1993). Predictive models for the breeder genetic algorithm: I. continous parameter optimization. *Evolutionary Computation*, *1*(1), 25–49.

Pelikan, M. (2002). *Bayesian optimization algorithm: From single level to hierarchy*. Doctoral dissertation, University of Illinois at Urbana-Champaign, Urbana, IL. (Also IlliGAL Report No. 2002023).

Pelikan, M., & Goldberg, D. E. (2001). Escaping hierarchical traps with competent genetic algorithms. *Proceedings of the Genetic and Evolutionary Computation Conference*, 511–518. (Also IlliGAL Report No. 2000020).

Pelikan, M., Goldberg, D. E., & Cantú-Paz, E. (2000a). Bayesian optimization algorithm, population sizing, and time to convergence. *Proceedings of the Genetic and Evolutionary Computation Conference*, 275–282. (Also IlliGAL Report No. 2000001).

Pelikan, M., Goldberg, D. E., & Cantú-Paz, E. (2000b). Linkage learning, estimation distribution, and Bayesian networks. *Evolutionary Computation*, *8*(3), 314–341. (Also IlliGAL Report No. 98013).

Pelikan, M., Goldberg, D. E., & Sastry, K. (2001). Bayesian optimization algorithm, decision graphs, and Occam's razor. *Proceedings of the Genetic and Evolutionary Computation Conference*, 519–526. (Also IlliGAL Report No. 2000020).

Pelikan, M., Lobo, F., & Goldberg, D. E. (2002). A survey of optimization by building and using probabilistic models. *Computational Optimization and Applications*, *21*, 5–20. (Also IlliGAL Report No. 99018).

Pelikan, M., & Sastry, K. (2004, January). *Fitness inheritance in the bayesian optimization algorithm* (IlliGAL Report No. 2004009). Urbana, IL: University of Illinois at Urbana-Champaign.

Pelikan, M., Sastry, K., & Goldberg, D. E. (2003). Scalability of the Bayesian optimization algorithm. *International Journal of Approximate Reasoning*, *31*(3), 221–258. (Also IlliGAL Report No. 2001029).

Sastry, K. (2001). *Evaluation-relaxation schemes for genetic and evolutionary algorithms*. Master's thesis, University of Illinois at Urbana-Champaign, General Engineering Department, Urbana, IL. (Also IlliGAL Report No. 2002004).

Sastry, K., & Goldberg, D. E. (2000). On extended compact genetic algorithm. *Late-Breaking Paper at the Genetic and Evolutionary Computation Conference*, 352–359. (Also IlliGAL Report No. 2000026).

Sastry, K., & Goldberg, D. E. (2002). Genetic algorithms, efficiency enhancement, and deciding well between fitness function with differing variances. *Proceedings of the Genetic and Evolutionary Computation Conference*, 528–535. (Also IlliGAL Report No. 2002002).

Sastry, K., Goldberg, D. E., & Pelikan, M. (2001). Don't evaluate, inherit. *Proceedings of the Genetic and Evolutionary Computation Conference*, 551–558. (Also IlliGAL Report No. 2001013).

Smith, R., Dike, B., & Stegmann, S. (1995). Fitness inheritance in genetic algorithms. In *Proceedings of the ACM Symposium on Applied Computing* (pp. 345–350). New York, NY, USA: ACM.

Takagi, H. (2001). Interactive evolutionary computation: Fusion of the capabilities of EC optimization and human evaluation. *Proceedings of the IEEE*, *89*(9), 1275–1296.

Thierens, D., & Goldberg, D. E. (1994). Convergence models of genetic algorithm selection schemes. *Parallel Problem Solving from Nature*, *3*, 116–121.

Zheng, X., Julstrom, B., & Cheng, W. (1997). Design of vector quantization codebooks using a genetic algorithm. *Proceedings of the IEEE Conference on Evolutionary Computation, ICEC*, 525–529.

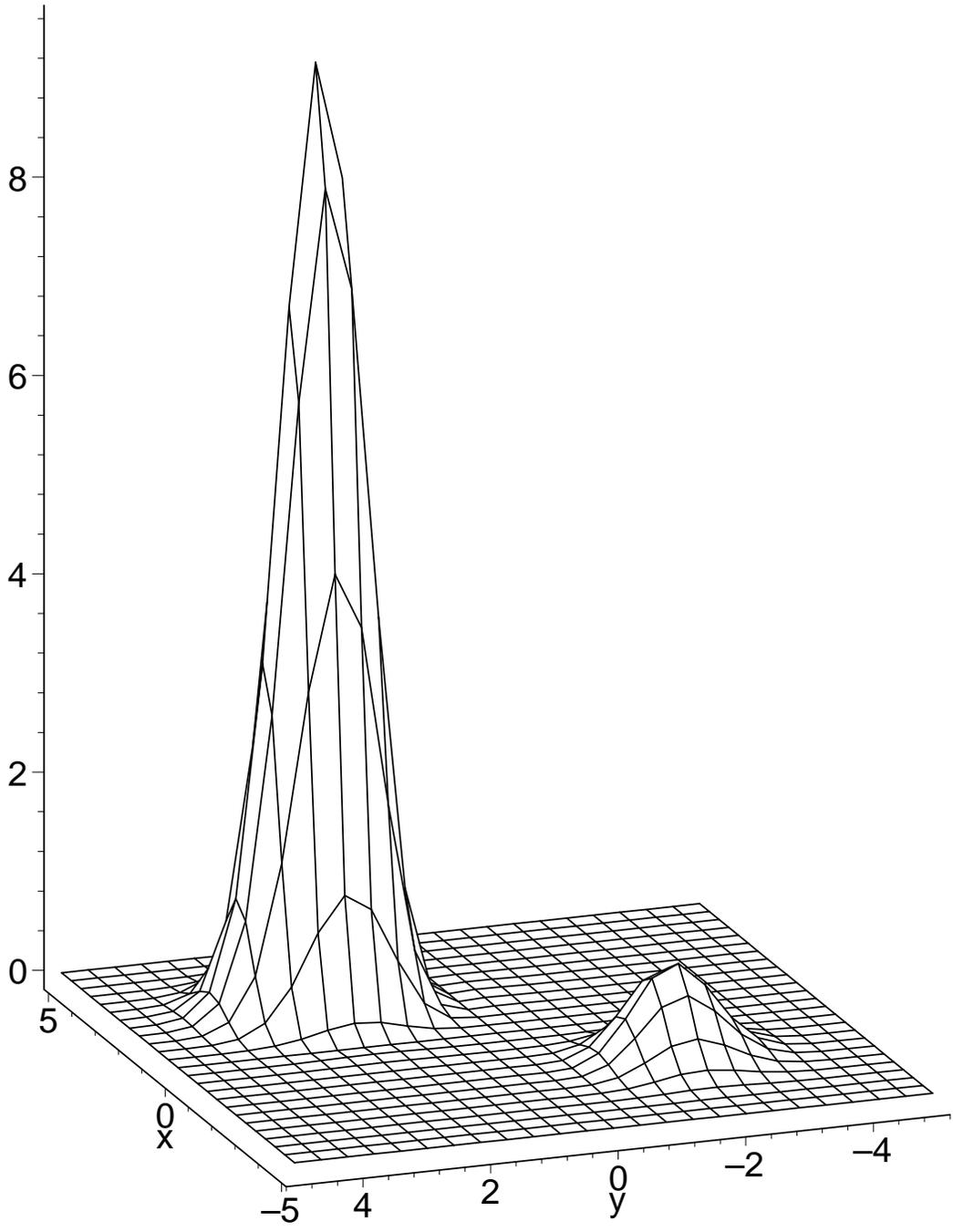